%% file: main.tex
\pdfoutput=1

\documentclass[11pt]{article}

\usepackage[final]{acl}

\usepackage{times}
\usepackage{latexsym}
 
\usepackage{amsmath}
\usepackage{amssymb}

\usepackage{graphicx}
\graphicspath{{images/}} 

\usepackage{array}
\usepackage{multirow}

\usepackage{subcaption}

\usepackage[T1]{fontenc}

\usepackage[utf8]{inputenc}

\usepackage{microtype}

\usepackage{inconsolata}

\usepackage{graphicx}

\usepackage{booktabs}

\usepackage{xspace}

\newcommand{\pearfamily}{\textsc{PEAR}\xspace}
\newcommand{\pearfamilybf}{\textbf{PEAR}\xspace}

\newcommand{\pear}{\textsc{PEAR}\xspace}          
\newcommand{\pearxl}{\textsc{PEAR-XL}\xspace}     

\newcommand{\pearboth}{\textsc{PEAR}$_{\mathrm{both}}$\xspace}
\newcommand{\pearbothkd}{\textsc{PEAR}$_{\mathrm{both,KD}}$\xspace}

\title{{\pearfamily{}}: {P}airwise {E}valuation for {A}utomatic {R}elative Scoring \\in Machine Translation}

\author{
  Lorenzo Proietti$^{1}$\thanks{Work conducted during an internship at Microsoft.} \quad\quad
  Roman Grundkiewicz$^{2}$ \quad\quad
  Matt Post$^{2}$ \\
  $^{1}$Sapienza University of Rome \quad
  $^{2}$Microsoft \\
  \texttt{lproietti@diag.uniroma1.it} \quad
  \texttt{\{rogrundk,mattpost\}@microsoft.com}
}

\begin{document}
\maketitle
\begin{abstract}
We present PEAR (Pairwise Evaluation for Automatic Relative Scoring), a supervised quality estimation (QE) metric family that reframes reference-free machine translation (MT) evaluation as a graded pairwise comparison. Given a source segment and two candidate translations, PEAR predicts the direction and magnitude of their quality difference. The metrics are trained using pairwise supervision derived from differences in human judgments, with an additional regularization term that encourages sign inversion under candidate order reversal.
On the WMT24 meta-evaluation benchmark, PEAR outperforms strictly matched single-candidate QE baselines trained with the same data and backbones, isolating the benefit of the proposed pairwise formulation. Despite using substantially fewer parameters than recent large metrics, PEAR surpasses far larger QE models and reference-based metrics. Our analysis further indicates that PEAR yields a less redundant evaluation signal relative to other top metrics. 
Finally, we show that PEAR is an effective utility function for minimum Bayes risk (MBR) decoding, reducing pairwise scoring cost at negligible impact.
\end{abstract}

\section{Introduction}
\label{sec:intro}

Automatic metrics are a primary tool for comparing modern machine translation (MT) systems. Shared task evaluations and much of the research literature rely heavily on metric scores \citep{marie-etal-2021-scientific,kocmi-etal-2021-ship,kocmi-etal-2024-navigating,kocmi-etal-2024-findings,kocmi-etal-2025-findings,kocmi-etal-2025-findings-wmt25}. Human evaluation remains the highest-quality signal, but it is expensive and difficult to scale across the growing number of systems, domains, and language pairs of interest \citep{zouhar-etal-2025-ai}. As MT quality improves, differences between strong systems become subtle, making both human and automatic discrimination harder \citep{proietti-etal-2025-machine,proietti-etal-2025-estimating,zouhar2025selectdatapointsefficienthuman}.

\input{figures/pear_draw}

Despite their diversity, most MT evaluation metrics, including quality estimation (QE) metrics that do not require references, share a structural property: they evaluate one candidate translation at a time and output an absolute scalar score. Typically, a metric conditions on the source segment, an optional reference translation, and a single candidate translation, and outputs a scalar score. We posit that this single-candidate perspective is mismatched to several aspects of modern MT evaluation, especially in the high-quality settings where differences are subtle. In addition, one of the dominant downstream use cases is comparative, including model selection and candidate translations ranking \citep{kocmi-etal-2021-ship,kocmi-etal-2024-navigating,perrella-etal-2024-beyond}. Motivated by evidence that comparative human judgments can be more consistent than absolute ratings in MT and related settings \citep{karpinska-etal-2021-perils,song-etal-2025-enhancing}, we ask whether supervised QE for comparing translations can be better framed as a graded pairwise comparison task.

We present \pearfamilybf{} (\textbf{P}airwise \textbf{E}valuation for \textbf{A}utomatic \textbf{R}elative Scoring), a novel supervised QE metric family that reframes reference-free MT evaluation as graded pairwise comparison. Given a source segment and two candidate translations, \pearfamily{} predicts the direction and magnitude of their quality difference. The metrics are trained on pairwise supervision constructed from differences in human segment-level judgments, with an additional regularization term that encourages sign inversion under candidate order reversal, yielding a relative score as output.

Our main contributions are as follows:
\begin{itemize}
\setlength{\itemsep}{1pt}
    \item We introduce \pearfamily{}, a novel supervised QE formulation for MT evaluation based on graded pairwise relative scoring.
    \item We provide controlled comparisons showing that the proposed pairwise formulation outperforms matched single-candidate QE under the same training setup, and that \pearfamily{} also surpasses much larger QE models and reference-based metrics.
    \item We show that \pearfamily{} avoids the need for quadratic system-to-system comparisons via a reference-anchored inference mode that does not rely on human references, remaining effective even when anchored to an MT output.
    \item We examine how \pearfamily{} correlates with other metrics, providing evidence of reduced redundancy.
    \item We demonstrate \pearfamily{}'s effectiveness and efficiency as a utility function for minimum Bayes risk (MBR) decoding.
\end{itemize}

Figure~\ref{fig:pear-draw} illustrates the \pearfamily{} execution flow at both the segment and system levels. We release code and trained checkpoints of \pearfamily{} metrics.\footnote{\url{https://github.com/prosho-97/pear}}

\section{Background and Related Work}
\label{sec:related}

\paragraph{Automatic MT evaluation metrics.}
Early MT metrics estimate quality via surface overlap against a reference translation, with BLEU \citep{papineni-etal-2002-bleu} and chrF \citep{popovic-2015-chrf} as widely used examples. Learned metrics based on pretrained representations improve agreement with human judgments by fine-tuning on human annotation signals, including reference-based models such as COMET and BLEURT \citep{rei-etal-2020-comet,sellam-etal-2020-bleurt,sellam-etal-2020-learning} and more recent large-scale approaches such as XCOMET and MetricX \citep{guerreiro-etal-2024-xcomet,juraska-etal-2024-metricx}. 
In parallel, QE metrics remove the dependency on references and score translations conditioned on the source and candidate only, including CometKiwi and its larger variants, as well as QE models of more recent learned metrics \citep{rei-etal-2022-cometkiwi,rei-etal-2023-scaling,guerreiro-etal-2024-xcomet,juraska-etal-2024-metricx}. Judging approaches based on large language models, including GEMBA prompting for MQM or ESA, have shown strong performance but are often expensive and may raise reproducibility concerns \citep{kocmi-federmann-2023-large,kocmi-federmann-2023-gemba}.

\paragraph{Pairwise and preference-based formulations.}
Pairwise evaluation has a long history in MT, including ranking-based training with engineered features and structured models \citep{ye-etal-2007-sentence,duh-2008-ranking,guzman-etal-2014-learning,guzman-etal-2015-pairwise}. Several learned metrics leverage comparative supervision during training while still producing single-candidate scores at inference time, such as COMET-RANK \citep{rei-etal-2020-comet} and reward-modeling approaches based on pairwise preferences \citep{tan-monz-2025-remedy}. 
By comparison, COMET-poly incorporates additional context beyond the single translation at inference time, grounding the evaluation of a candidate in other candidates for the same source \citep{zufle-etal-2025-comet}. Closer to our inference setup, MT-RANKER formulates reference-free MT evaluation as binary classification: given a source and two candidates, it predicts which translation is better; consequently, its output space does not represent ties or the strength of the preference \citep{moosa2024mtranker}. \pearfamily{} differs by predicting graded relative differences without ruling out ties by design, and encouraging sign inversion under candidate order reversal.

\paragraph{Comparative human judgments.}
Ranking-based protocols have long been used in MT \citep{callison-burch-etal-2008-meta,bojar-etal-2016-findings,sakaguchi-van-durme-2018-efficient}, and recent human evaluations increasingly adopt comparative setups, such as side-by-side MQM, motivated in part by improved consistency and agreement when annotators compare two candidates directly \citep{song-etal-2025-enhancing}. The same trend is also reflected in recent platforms for human evaluation in MT that support multi-candidate assessment \citep{zouhar2026pearmuthumanevaluationtranslation}. More generally, this mirrors broader observations that comparative judgments reduce subjectivity in open-ended generation evaluation \citep{karpinska-etal-2021-perils}. \pearfamily{} draws on this motivation but targets the automatic setting, treating comparative scoring as the prediction target for supervised QE in MT.

\section{\pearfamily{}: Pairwise Evaluation for Automatic Relative Scoring}
\label{sec:pear}

In this section, we introduce \pearfamily{}. 
We begin by briefly recalling the standard regression-based formulation, where a metric assigns an absolute quality score to a single candidate translation. We then describe \pearfamily{}, which instead predicts a graded quality difference between two candidate translations of the same source text, together with the model architecture, training objective, and inference procedure.

\subsection{From Absolute to Relative Scoring}
\label{subsec:pear-absolute-vs-relative}

Supervised QE metrics are trained to predict an absolute score from a source segment $s$ and a single candidate translation $mt$. In reference-based evaluation, the metric is additionally provided with a reference translation $r$. Accordingly, these metrics take one of the following two forms:
\begin{equation*}
\begin{aligned}
\hat{s}(s,mt)   &= g_\theta(s,mt), \\
\hat{s}(s,mt,r) &= g_\theta(s,mt,r).
\end{aligned}
\end{equation*}
They are typically optimized to fit human segment-level judgments $s_h(s,mt)$\footnote{Several recent human evaluation protocols for MT are reference-free: annotators assess system outputs only with respect to the source text, which reduces dependence on any particular reference and thereby mitigates reference bias \citep{freitag-etal-2021-experts,kocmi-etal-2022-findings,kocmi-etal-2024-error}.} via a regression loss \citep{sellam-etal-2020-bleurt,rei-etal-2020-comet,rei-etal-2022-cometkiwi,guerreiro-etal-2024-xcomet,juraska-etal-2024-metricx};\footnote{In practice, this regression loss is commonly implemented as Mean Squared Error (MSE) between the predicted scores and the human judgments.} other work casts automatic MT evaluation as reward modeling, learning from human preferences while still producing absolute scores for individual candidate translations at inference time \citep{tan-monz-2025-remedy}.
When comparing two candidates $mt_a$ and $mt_b$, an implicit relative score can then be obtained by subtraction, i.e., by computing $\hat{s}(s,mt_a)-\hat{s}(s,mt_b)$ in reference-less evaluation, or $\hat{s}(s,mt_a,r)-\hat{s}(s,mt_b,r)$ in reference-based evaluation.

Rather than deriving a comparison by subtracting two independently predicted absolute scores, an alternative approach is to predict the preference directly from a joint encoding of the source and the two candidate translations. MT-RANKER follows this approach in the QE setting, taking $(s,mt_a,mt_b)$ as input and predicting which candidate translation is better through a binary classification framing \citep{moosa2024mtranker}. However, this formulation collapses comparison to a two-way decision: it cannot quantify the strength of the preference, and it rules out ties by design, since they are not part of the output label space. \pearfamily{} instead treats graded comparison as the target of prediction. Given a source text, it scores a pair of candidate translations in a shared input context, producing a signed scalar preference.

\subsection{Task Formulation}
\label{subsec:pear-formulation}

Let $s$ be a source segment and let $mt_a, mt_b$ be two candidate translations of $s$. A \pearfamily{} model $f_\theta$ predicts a real-valued relative score
\begin{equation*}
\hat{\Delta}_{ab} \;=\; f_\theta(s, mt_a, mt_b) \in \mathbb{R},
\end{equation*}
interpreted as the predicted quality difference between $mt_a$ and $mt_b$ as translations of $s$. Positive values favor $mt_a$, negative values favor $mt_b$, and the closer the score is to zero, the more similar the predicted quality of the two candidates becomes, with the preference approaching a tie.

\pearfamily{} is trained to approximate human pairwise differences. Specifically, given human segment-level scores $s_h(s,mt)$, we construct supervision for a candidate pair as:
\begin{equation*}
\Delta^\star_{ab} \;=\; s_h(s,mt_a) \;-\; s_h(s,mt_b).
\end{equation*}
At the system level, let $\{s_i\}_{i=1}^n$ be a test set, and let $\{(mt_i^A, mt_i^B)\}_{i=1}^n$ be the corresponding output pairs produced by two MT systems $S_A$ and $S_B$. \pearfamily{} produces segment-level predictions $\hat{\Delta}_i = f_\theta(s_i, mt_i^A, mt_i^B)$, which are aggregated by arithmetic mean to yield a system-level graded preference:
\begin{equation*}
\hat{\Delta}(S_A,S_B) \;=\; \frac{1}{n}\sum_{i=1}^n \hat{\Delta}_i,
\end{equation*}
as also illustrated in Figure~\ref{fig:pear-draw}.

\subsection{Model Architecture and Input Formatting}
\label{subsec:pear-arch}

\pearfamily{} adopts a cross-encoder setup that jointly encodes the source text and the two candidate translations. We instantiate the encoder with InfoXLM Large\footnote{\url{https://huggingface.co/microsoft/infoxlm-large}} for \pear{} \citep{chi-etal-2021-infoxlm} and with XLM-RoBERTa-XL\footnote{\url{https://huggingface.co/facebook/xlm-roberta-xl}} for \pearxl{} \citep{goyal-etal-2021-larger}. Throughout, we describe the scoring computation for a single input sample, from input construction to the final relative score.

\paragraph{Input serialization.}
Given a source segment $s$ and two candidate translations $mt_a$ and $mt_b$, we serialize them into a single sequence:
\begin{equation*}
x \;=\; [\texttt{BOS}] \; s \; [\texttt{SEP}] \; mt_a \; [\texttt{SEP}] \; mt_b \; [\texttt{EOS}],
\end{equation*}
where \texttt{[BOS]}, \texttt{[SEP]}, and \texttt{[EOS]} denote the encoder's special tokens.\footnote{For InfoXLM and XLM-R encoders, these correspond to \texttt{<s>}, \texttt{</s></s>}, and \texttt{</s>}, respectively.}
We additionally build binary span masks $\mathbf{m}_{\mathrm{src}}, \mathbf{m}_{a}, \mathbf{m}_{b} \in \{0,1\}^{T}$ over token positions in $x$ that select the content tokens of $s$, $mt_a$, and $mt_b$.\footnote{Special tokens are masked out in this process.}

\paragraph{Encoder representations.}
We denote the contextual token representations extracted from the encoder's final layer by:
\begin{equation*}
\mathbf{H} = \mathrm{Enc}_\phi(x) \in \mathbb{R}^{T \times d},
\end{equation*}
where $T$ denotes the number of tokens in $x$ and $d$ is the encoder hidden dimensionality. We apply masked mean pooling to obtain span representations:
\begin{equation*}
\mathbf{h}_{\ell}
\;=\;
\frac{\sum_{t=1}^{T} m_{\ell,t}\,\mathbf{H}_{t}}
{\max\!\left(\sum_{t=1}^{T} m_{\ell,t},\, \varepsilon\right)}
\quad \text{for } \ell \in \{\mathrm{src}, a, b\},
\end{equation*}
where $\mathbf{H}_{t} \in \mathbb{R}^{d}$ is the token vector at position $t$ and $\varepsilon>0$ avoids division by zero.

\paragraph{Pairwise head.}
We construct a source-aware representation for each candidate translation $k \in \{a,b\}$:
\begin{equation*}
\boldsymbol{\varphi}_{k} = \bigl[\mathbf{h}_{k};\; \mathbf{h}_{k}\odot \mathbf{h}_{\mathrm{src}};\; |\mathbf{h}_{k}-\mathbf{h}_{\mathrm{src}}|\bigr] \in \mathbb{R}^{3d},
\end{equation*}
where $\odot$ and $|\cdot|$ are applied elementwise. Using shared parameters, we map $\boldsymbol{\varphi}_{k}$ to a scalar utility term:
\begin{align*}
\mathbf{a}_{k} &= \mathrm{Proj}(\boldsymbol{\varphi}_{k}) \in \mathbb{R}^{d}, \\
u_{k} &= \mathrm{FFN}(\mathbf{a}_{k}) \in \mathbb{R}.
\end{align*}
$\mathrm{Proj}$ is a linear projection followed by GELU \cite{hendrycks-gimpel-2016-gelu}, layer normalization \cite{ba2016layernormalization}, and dropout, while $\mathrm{FFN}$ is a multilayer feed-forward head with three hidden linear layers,\footnote{For PEAR, the hidden sizes are $512 \rightarrow 256 \rightarrow 128$; for PEAR-XL, they are $2048 \rightarrow 1024 \rightarrow 512$.} each followed by GELU and dropout. We then form a comparison logit by subtraction:
\begin{equation*}
z \;=\; u_{a} - u_{b}.
\end{equation*}
Note that only the difference is used for supervision and inference, so the individual $u_k$ values are not intended as absolute quality scores. Then, the final output is a scaled relative score:
\begin{equation*}
\hat{\Delta}_{ab} \;=\; \alpha\, z,
\qquad
\alpha \;=\; \mathrm{softplus}(\alpha_{\mathrm{raw}}) + \varepsilon,
\end{equation*}
where $\alpha_{\mathrm{raw}}\in\mathbb{R}$ is a learned scalar parameter. The softplus reparameterization, together with the addition of a small constant $\varepsilon>0$, ensures $\alpha>0$.

\subsection{Training Objective}
\label{subsec:pear-objective}

A key property of pairwise relative scoring is sign inversion under candidate order reversal: reversing the candidate order should negate the predicted difference while preserving its magnitude, i.e., $f_\theta(s,mt_a,mt_b) = -f_\theta(s,mt_b,mt_a)$. To encourage this behavior, for each training instance $(s,mt_a,mt_b)$ we also consider the reversed order $(s,mt_b,mt_a)$ and denote the corresponding prediction by $\hat{\Delta}_{ba}=f_\theta(s,mt_b,mt_a)$. Specifically, we train with a Huber loss \citep{huber1964robust} on the human difference together with an antisymmetry regularization term:
\begin{align*}
\mathcal{L}_{\text{diff}} &= \ell_{\delta}\!\left(\hat{\Delta}_{ab} - \Delta^\star_{ab}\right), \\
\mathcal{L}_{\text{flip}} &= \bigl(\hat{\Delta}_{ab} + \hat{\Delta}_{ba}\bigr)^2, \\
\mathcal{L} &= \mathcal{L}_{\text{diff}} + \lambda_{\text{flip}} \,\mathcal{L}_{\text{flip}},
\end{align*}
where $\lambda_{\text{flip}}$ is a hyperparameter controlling the strength of the antisymmetry objective.\footnote{Appendix~\ref{app:pear_consistency} analyzes the impact of this additional training objective on antisymmetry and additive transitivity in \pear{}.} Here $\ell_{\delta}$ denotes the Huber loss:
\begin{equation*}
\ell_{\delta}(r) \;=\;
\begin{cases}
\frac{1}{2}r^2 & \text{if } |r|\le \delta,\\
\delta\left(|r|-\frac{1}{2}\delta\right) & \text{otherwise,}
\end{cases}
\end{equation*}
where $\delta>0$ is a hyperparameter that sets the residual magnitude at which the penalty transitions from quadratic to linear.\footnote{Compared to MSE, the Huber loss is less sensitive to large residuals \citep{huber1964robust}; in Appendix~\ref{app:huber_vs_mse}, we report an ablation where we show that it improves performance over MSE in our setting.}

\subsection{Inference Modes}
\label{subsec:pear-inference}

We use \pearfamily{} in two inference configurations. In its default pairwise QE mode, it compares two candidate translations of the same source segment without access to reference translations. When a human reference is available, we additionally consider a reference-anchored mode by fixing one side of the pair to the reference.

\paragraph{Pairwise QE mode (\pearfamily{}).}
Given a source segment $s$ and two candidate translations $mt_a$ and $mt_b$, \pearfamily{} returns a relative score:
\begin{equation*}
\hat{\Delta}_{ab} \;=\; f_{\theta}(s, mt_a, mt_b).
\end{equation*}
To further reduce sensitivity to the candidates' order, we optionally also score the swapped order and combine the two predictions:
\begin{align*}
\hat{\Delta}_{ba} \;&=\; f_{\theta}(s, mt_b, mt_a), \\
\tilde{\Delta}_{ab} \;&=\; \tfrac{1}{2}\bigl(\hat{\Delta}_{ab}-\hat{\Delta}_{ba}\bigr).
\end{align*}
This bidirectional variant (denoted \pearboth) averages the two relative scores obtained from both input orders.

\paragraph{Reference-anchored mode (\pearfamily{}$_{\mathrm{ref}}$).}
When a human reference translation $r$ is available, we anchor the comparison by fixing one side of the input to the reference:
\begin{equation*}
\hat{\Delta}(mt, r) \;=\; f_{\theta}(s, mt, r).
\end{equation*}
This does not make \pearfamily{} a reference-based metric in the usual sense; rather, it instantiates the same relative-scoring function in a reference-anchored configuration. As above, we can also score the inverted order $f_{\theta}(s, r, mt)$ and combine the two predictions with the same rule.

A practical advantage of this reference-anchored configuration is computational. When evaluating $N$ systems on an evaluation set $\{(s_i, r_i)\}_{i=1}^{n}$ with system outputs $\{mt_i^{(j)}\}_{i=1}^{n}$ for each system $S_j$, it yields one reference-anchored score per system:
\begin{equation*}
\hat{\Delta}(S_j, r) \;=\; \frac{1}{n}\sum_{i=1}^{n} f_{\theta}(s_i, mt_i^{(j)}, r_i).
\end{equation*}
This requires $N$ system scores, avoiding the $N(N-1)/2$ system-to-system comparisons needed by fully pairwise evaluation.

\section{Experimental Setup}
\label{sec:setup}

We now summarize the empirical setting for our main experiments. We describe the training and evaluation procedures and report the trained \pearfamily{} models. Details about data, training hyperparameters, and baselines are reported in Appendix~\ref{subsec:baselines}.

\subsection{Training and Evaluation Data}
\label{subsec:data}

\paragraph{WMT supervision.}
We train \pearfamily{} on WMT human evaluation data, converting absolute segment-level human assessments into pairwise supervision via score differencing (Section~\ref{subsec:pear-formulation}).
Our training data combines DA, DA+SQM, and MQM annotations, with MQM offering the most fine-grained supervision.
Following common practices in MT metrics training, we adopt a two-stage schedule \citep{rei-etal-2022-comet, juraska-etal-2023-metricx, juraska-etal-2024-metricx, perrella-etal-2024-guardians, proietti-etal-2025-estimating}.
In the first training stage, we pre-train on DA and DA+SQM judgments released in the WMT evaluation campaigns from WMT16 to WMT23, providing broad coverage across language directions and translation quality.
In the second training stage, we fine-tune exclusively on MQM supervision from WMT20 to WMT23,\footnote{This is done not only because MQM offers the most fine-grained MT quality assessment, but also to align the training signal with our main target evaluation setting, since WMT24 meta-evaluation is centered on MQM.} additionally including the IndicMT Eval MQM dataset for English$\rightarrow$Indic directions \citep{sai-b-etal-2023-indicmt}, which has been used to train XCOMET \citep{guerreiro-etal-2024-xcomet}, a metric included in our comparisons.

\paragraph{Scaling via distilled MQM supervision.}
MQM gold data are available for relatively few language pairs and are expensive to collect.
To stress-test whether the proposed pairwise QE framing scales favorably with additional supervision, we also test \pearfamily{} models whose second training stage is augmented with MQM annotations distilled from GPT-4.1-mini \citep{openai2025gpt41} on language pairs not covered by MQM, prompting it with a GEMBA-MQM V2 approach \citep{junczys-dowmunt-2025-gemba}. Appendix~\ref{subsec:baselines} provides additional details on these data.

\paragraph{Evaluation benchmark.}
Our primary benchmark is the MQM test set released with the WMT24 Metrics Shared Task \citep{freitag-etal-2024-llms}. We adopt their official evaluation toolkit.\footnote{\url{https://github.com/google-research/mt-metrics-eval}}
Following the WMT24 setup, we use Soft Pairwise Accuracy (SPA) at the system level \citep{thompson-etal-2024-improving} and pairwise accuracy with tie calibration (acc$_{eq}^*$) at the segment level \citep{deutsch-etal-2023-ties}. 
Since \pearfamily{} produces pairwise scores by design, we integrate it with the toolkit by producing segment-level scores for each system pair.
We average SPA and acc$_{eq}^*$ over the three language pairs featuring human MQM annotations,\footnote{English--German, English--Spanish, and Japanese--Chinese.} and we also report average correlation (Avg Corr), defined as the mean of these two averages.

\subsection{Trained Models}
\label{subsec:impl-metaeval}

\paragraph{\pearfamily{} models.}
We train and evaluate two \pearfamily{} variants:
\textbf{\pearfamily{}}, instantiated with InfoXLM Large \citep{chi-etal-2021-infoxlm},
and \textbf{\pearfamily{}-XL}, instantiated with XLM-RoBERTa-XL \citep{goyal-etal-2021-larger}.

\paragraph{Matched absolute-scoring baselines.}
To disentangle the effect of the proposed pairwise QE formulation from backbone capacity and training data exposure, we also train matched absolute-scoring QE baselines (\textbf{Single-QE} and \textbf{Single-QE-XL}) that take only the source segment and a single candidate translation as input, share the same encoder backbones and training data, and predict an absolute scalar quality score for that candidate. These baselines let us validate the proposed pairwise formulation at the methodological level, testing it as an alternative to conventional absolute-score QE under matched training conditions.

\input{tables/pear_vs_single_1col}

\input{tables/wmt24_main_res}

\section{Results and Analyses}
\label{sec:res}

This section reports empirical evidence for \pearfamily{} across several settings. We start with reporting results on the WMT24 Metrics Shared Task benchmark, and continue by showing additional analyses.

\subsection{Pairwise vs.\ Single-Candidate QE at Scale}
\label{subsec:scale}

We separate the effect of the proposed pairwise QE formulation from confounding factors such as backbone capacity, training data exposure, and model selection comparing \pearfamily{} against strictly matched single-candidate QE baselines trained with the same data, hyperparameters, and backbone model. Checkpoint selection is based on held-out WMT23 MQM data,\footnote{The WMT23 MQM dataset includes English--German, Chinese--English, and Hebrew--English.} and results are reported on the WMT24 MQM benchmark.

We report results for \pearfamily{} in its default configuration (one input order) and for the bidirectional variant that combines predictions from both input orders, matching the inference mode described in Section~\ref{subsec:pear-inference}. We further consider settings with and without knowledge distillation augmentation in the second training stage, to probe whether the advantage of our pairwise QE formulation persists---and potentially widens---as the amount of MQM supervision is scaled up via distilled annotations.

Table~\ref{tab:pear_vs_single} reports the comparison results. \pearfamily{} yields better performance than the matched single-candidate baselines, indicating that the proposed pairwise QE framing improves over single-candidate QE for comparing candidate translations. The statistical significance analysis further supports this conclusion: each PEAR model outperforms its corresponding Single-QE counterpart with a better statistical significance rank. In addition, the near-identical performance of the single-order and \textit{both} variants suggests that \pearfamily{} largely satisfies sign inversion under candidate order reversal.

\subsection{Comparison with MT-RANKER}
MT-RANKER is the closest prior approach to \pearfamily{} in terms of formulation, but it predicts a strict pairwise preference and does not model ties \citep{moosa2024mtranker}. To enable a fair comparison, we repurpose the WMT24 MQM test set into a binary classification test set by removing all candidate translation pairs whose human MQM assessments are tied. This yields 49{,}666 pairs for English--German, 24{,}777 for English--Spanish, and 34{,}045 for Japanese--Chinese. We then compare PEAR and PEAR$_{\mathrm{KD}}$ in their default single-order configuration against the best-performing MT-RANKER model, i.e., MT-RANKER-XXL. 
Table~\ref{tab:mtranker_comparison} reports segment-level pairwise accuracy on non-tied pairs across the three WMT24 language pairs featuring human MQM annotations, together with macro-average. Both \pearfamily{} models outperform MT-RANKER-XXL while using substantially fewer parameters.

\input{tables/mtranker_comparison}

\subsection{WMT24 Results}
\label{subsec:wmt24_res}

Table~\ref{tab:wmt24_main_res} reports results on the WMT24 meta-evaluation benchmark. For \pearfamily{}, we report only the bidirectional configuration (\textit{both}), since Table~\ref{tab:pear_vs_single} shows that single-pass and \textit{both} yield very similar performance. All \pearfamily{} variants in Table~\ref{tab:wmt24_main_res} are trained with the same hyperparameters used in the controlled comparison with the only difference that we also include WMT23 data in training.

\paragraph{Better performance with fewer parameters}
Among reference-free metrics, \pearfamily{} attains high average correlation with substantially smaller models. \pearfamily{}-XL$_{\mathrm{both}}$ (3.5B) exceeds MetricX-24-Hybrid-QE-XL (3.7B) and XCOMET-QE (24B) (70.2 vs.\ 69.9 and 69.5), while using roughly 7$\times$ fewer parameters than XCOMET-QE. Distilled supervision yields a further improvement, raising average correlation to 70.5.

The same trend holds with the smaller \pearfamily{} models. \pearboth (560M) remains higher than MetricX-24-Hybrid-QE-XL (3.7B) and XCOMET-QE (24B) on average correlation (70.1 vs.\ 69.9 and 69.5), despite using about 7$\times$ and 40$\times$ fewer parameters, respectively. The only QE metric comparable in size to \pearfamily{} is CometKiwi (560M), yet it yields a markedly lower result (64.0 vs.\ 70.1). With distilled supervision, \pearbothkd also edges out CometKiwi-XXL (70.4 vs.\ 70.3), while being about 20$\times$ smaller (560M vs.\ 10.5B).

\paragraph{Comparison against reference-based metrics}
Even without references, \pearfamily{} compares favorably to reference-based metrics. For example, \pearfamily{}-XL$_{\mathrm{both,KD}}$ matches average correlation of MetricX-24-Hybrid-Large (70.5 vs.\ 70.5), and \pearboth exceeds COMET-22 and BLEURT-20 (70.1 vs.\ 68.9 and 68.6).

\input{figures/pearref_anchor_ranks}

\subsection{Does \pearfamily{}$_{\mathrm{ref}}$ Require References?}
\label{subsec:pearref_refs}

Table~\ref{tab:wmt24_main_res} shows that the reference-anchored configuration, \pearfamily{}$_{\mathrm{ref}}$, slightly outperforms its corresponding QE (\textit{both}) variant on WMT24. We now test whether these performances depend on anchoring to a human reference, or whether the same behavior holds when the anchor is an MT output.

We instantiate the anchor slot with (i) the human reference (\textit{Human Ref}) and (ii) the output of six MT systems. For each MT anchor, we re-run WMT24 meta-evaluation after removing the anchor system from the benchmark, since its outputs are used as the fixed comparison target. Among the six MT anchors, we also include MSLC as a more extreme stress test: in the English--German MQM human evaluation reported by \citet{freitag-etal-2024-llms}, for example, Llama3-70B has a system-level MQM score of -3.62, whereas MSLC obtains -13.46, making it the lowest-quality system by a large margin.

Figure~\ref{fig:pearref_anchor_ranks} shows that the ranks\footnote{Since the set of evaluated systems changes with the chosen MT anchor, the resulting Avg Corr values are not directly comparable across anchors.} of \pearfamily{}$_{\mathrm{ref}}$ variants are stable across anchors. This picture remains largely unchanged even when \pearfamily{}$_{\mathrm{ref}}$ is anchored to MSLC, a substantially lower-quality MT system according to the MQM human evaluation results of \citet{freitag-etal-2024-llms}. This suggests that \pearfamily{}$_{\mathrm{ref}}$ does not rely on human references to retain its performance.

\subsection{Segment-level Correlation Between \pearfamily{} and Other Metrics}
\label{subsec:pear_metric_corr}

\pearfamily{} is trained with a pairwise relative-scoring objective, unlike most supervised metrics that use single-candidate regression.
To assess how distinct \pearfamily{}’s evaluation signal is, we analyze its segment-level correlation with other metrics.

For this analysis, we compute the Pearson correlation over pairwise segment-level difference scores. For \pearfamily{}, we use its predicted pairwise scores directly. For single-candidate metrics $m$ that assign a scalar score to each candidate translation (i.e., $m(s,mt)\in\mathbb{R}$), we obtain pairwise differences by subtracting their segment-level scores for each MT system pair: $\Delta_m(s,mt_a,mt_b)=m(s,mt_a)-m(s,mt_b)$.\footnote{For simplicity, we use reference-free notation here. In the correlation analysis, we also include reference-based metrics that additionally condition on a human reference.} We compute correlations separately for each WMT24 MQM language pair.

Figures~\ref{fig:pear_corr_en_de}, \ref{fig:pear_corr_en_es}, and \ref{fig:pear_corr_ja_zh} in Appendix~\ref{app:corr_mats} show the correlation matrices for English--German, English--Spanish, and Japanese--Chinese. Across all language pairs, \pearboth and \pearbothkd exhibit consistently lower correlation with other top-performing WMT24 metrics compared to supervised and LLM-as-a-Judge approaches, suggesting that \pearfamily{} may provide a distinct evaluation signal rather than mirroring existing metrics. 
For example, the correlation of these two \pearfamily{} variants with MetricX-24-Hybrid-QE\footnote{It denotes the XXL checkpoint.} is moderate on English--German ($r\!\approx\!0.71$), drops on English--Spanish ($r\!\approx\!0.51$), and is much lower on Japanese--Chinese ($r\!\approx\!0.26$). 
Excluding lexical and unsupervised baselines, i.e., BLEU \citep{papineni-etal-2002-bleu}, chrF \citep{popovic-2015-chrf}, and BERTScore \citep{Zhang*2020BERTScore:}, \pearfamily{} metrics are the least correlated with the rest of the metric suite.
Importantly, this does not reflect weaker metric performance, but rather suggests a different evaluation behavior. 
Such lower correlation can be useful in MT fine-tuning, where diverse metrics help avoid over‑optimizing for a single signal.
We leave a deeper investigation of the sources and implications of these correlations to future work.

\input{tables/mbr}

\subsection{\pearfamily{} for MBR Decoding}
\label{subsec:pear_mbr}

Reference-based metrics such as COMET-22 \citep{rei-etal-2020-comet, freitag-etal-2024-llms} and BLEURT-20 \citep{sellam-etal-2020-learning} are trained to condition on human references, whereas in MBR decoding they are applied to candidate-to-candidate comparisons \citep{freitag-etal-2022-high}, creating a slight mismatch between training and use. \pearfamily{}, by contrast, is trained explicitly for reference‑free pairwise comparison, making it naturally suited for this setting. Here, we compare \pearfamily{}, COMET-22, and BLEURT‑20 as MBR utility functions.

\paragraph{Candidate lists.}
We translate the English source side of the WMT24 test set into German and Japanese with a transformer-based multilingual MT system trained on public and internal data with teacher-student knowledge distillation. We decode with beam search and retain a 100-best list for each source segment.
\paragraph{MBR utilities.}
We run MBR over each 100-best list using three utilities: \pearfamily{},\footnote{InfoXLM Large checkpoint fine-tuned without distilled supervision. We use the default inference configuration, not the \textit{both} variant.} and the reference-based metrics COMET-22 and BLEURT-20.\footnote{Their parameter counts are comparable to PEAR's.} For \pearfamily{}, we compare two implementations. The first computes the full utility matrix, while the second exploits antisymmetry and evaluates only one triangle of the matrix, using $u(h_j,h_i)=-u(h_i,h_j)$ to fill the other half, thereby halving the number of forward passes. This tests whether \pearfamily{} behaves as an approximately antisymmetric utility during decoding, while enabling a cheaper MBR procedure by reducing the number of required pairwise comparisons.

\paragraph{Results.}
We evaluate the resulting MBR outputs with three metrics: XCOMET-XL, CometKiwi-XL, and MetricX-24-Hybrid-XL. Table~\ref{tab:mbr_results} shows that \pearfamily{} yields nearly identical performance under the full and antisymmetry-reduced MBR implementations, indicating that the antisymmetry shortcut is effective in practice. Moreover, \pearfamily{}-based MBR improves over COMET-22 and BLEURT-20 under XCOMET-XL and MetricX-24-Hybrid-XL, while gains under CometKiwi-XL are relatively smaller. These results suggest that \pearfamily{}, trained explicitly for pairwise comparison, can be an effective utility for MBR decoding.

\section{Conclusion}
\label{sec:conclusion}

We presented \pearfamily{}, a supervised QE metric family that models MT evaluation as graded pairwise relative scoring. Unlike standard metrics that score one candidate at a time, \pearfamily{} compares two translations jointly and predicts both the direction and strength of preference, which mirrors the way metrics are often used in practice, namely to decide which MT system or candidate is better. In controlled experiments, \pearfamily{} consistently outperforms matched single-candidate QE baselines, indicating that the pairwise formulation is a more effective approach for comparing candidate translations.

On WMT24 MQM meta‑evaluation, \pearfamily{} achieves higher correlation than the largest QE submissions, despite its smaller model size. It further outperforms reference‑based metrics, despite not relying on human references.
We also showed that \pearfamily{} is an effective utility function for minimum Bayes risk (MBR) decoding, enabling a substantially cheaper pairwise scoring procedure with negligible impact on decoding quality.
We further found that \pearfamily{}$_{\mathrm{ref}}$ remains effective when the anchor translation is an MT output rather than a human reference, and that \pearfamily{}’s segment-level pairwise scores are less correlated with other top metrics, suggesting that \pearfamily{} captures a different evaluation signal.

Future work will focus on understanding which phenomena drive this divergence, and on leveraging pairwise relative scoring in other evaluation settings.

\section*{Limitations}
\label{sec:limitations}

\paragraph{Model scale.}
Our experiments do not fully characterize how \pearfamily{} performance scales with substantially larger models.
The largest \pearfamily{} checkpoint we fine-tune in this work is \pearxl (3.5B parameters), and we therefore do not test whether the performance gains we attribute to the pairwise formulation persist, widen, or saturate with larger models.
Recent works suggest that scaling up the underlying model and fine-tuning larger backbones can be a strong driver of performance for supervised metrics \citep{rei-etal-2023-scaling,guerreiro-etal-2024-xcomet,juraska-etal-2024-metricx,juraska-etal-2025-metricx,tan-monz-2025-remedy}.
Extending \pearfamily{} to larger model sizes is a natural next step.
In addition, the pairwise setup could be extended from scalar relative scoring to MQM sequence tagging that explicitly predicts error spans in each candidate translation \citep{perrella-etal-2022-matese,kocmi-federmann-2023-gemba,junczys-dowmunt-2025-gemba}, which is well aligned with side-by-side MQM protocols based on comparative judgment \citep{song-etal-2025-enhancing}.

\paragraph{Metric divergence.}
In Section~\ref{subsec:pear_metric_corr}, we report that \pearfamily{} exhibits lower correlation with other strong WMT24 metrics than is typical among top-performing metrics.
In this work, we treat this primarily as evidence that \pearfamily{} may induce a complementary evaluation signal, but we do not deeply investigate how this could be exploited.
For example, lower correlation could be beneficial when combining metrics in an ensemble, or when other metrics are used as optimization targets, such as rewards in reinforcement learning, since it may reduce over-optimization to a single signal and support fairer automatic assessment, but it could also indicate sensitivity to specific phenomena that warrant targeted diagnosis. A more detailed analysis of this divergence is left for future work.

\paragraph{Targeted synthetic data.}
Unlike some recent metrics that incorporate targeted synthetic examples designed to address known failure modes (e.g., fluent but unrelated translations, undertranslation, and related specific errors), \pearfamily{} is not fine-tuned with comparable hand-designed perturbation suites \citep{juraska-etal-2023-metricx,juraska-etal-2024-metricx,guerreiro-etal-2024-xcomet}.
Although we experiment with scaling supervision via distilled MQM annotations, we do not study whether \pearfamily{} particularly benefits from synthetic pair construction that is explicitly contrastive by design.
\pearfamily{}’s pairwise formulation makes targeted contrastive pair construction more natural and potentially more cost-effective than for other traditional supervised metrics.
We leave a systematic investigation of such targeted synthetic data strategies, and how they interact with training on graded pairwise comparisons, for future work.

\section*{Ethics Statement}
We foresee no ethical issues with our work.

\section*{Acknowledgements}
This work was carried out while Lorenzo Proietti was enrolled in the Italian National Doctorate on Artificial Intelligence run by Sapienza University of Rome. The authors thank Marcin Junczys-Dowmunt for his valuable help with the knowledge distillation process from GPT-4.1-mini.

\bibliography{anthology-1, anthology-2, custom}

\appendix
\input{appendix.tex}

\end{document}

%% file: figures/pear_draw.tex
\begin{figure}[t]
  \centering
  \includegraphics[width=\columnwidth]{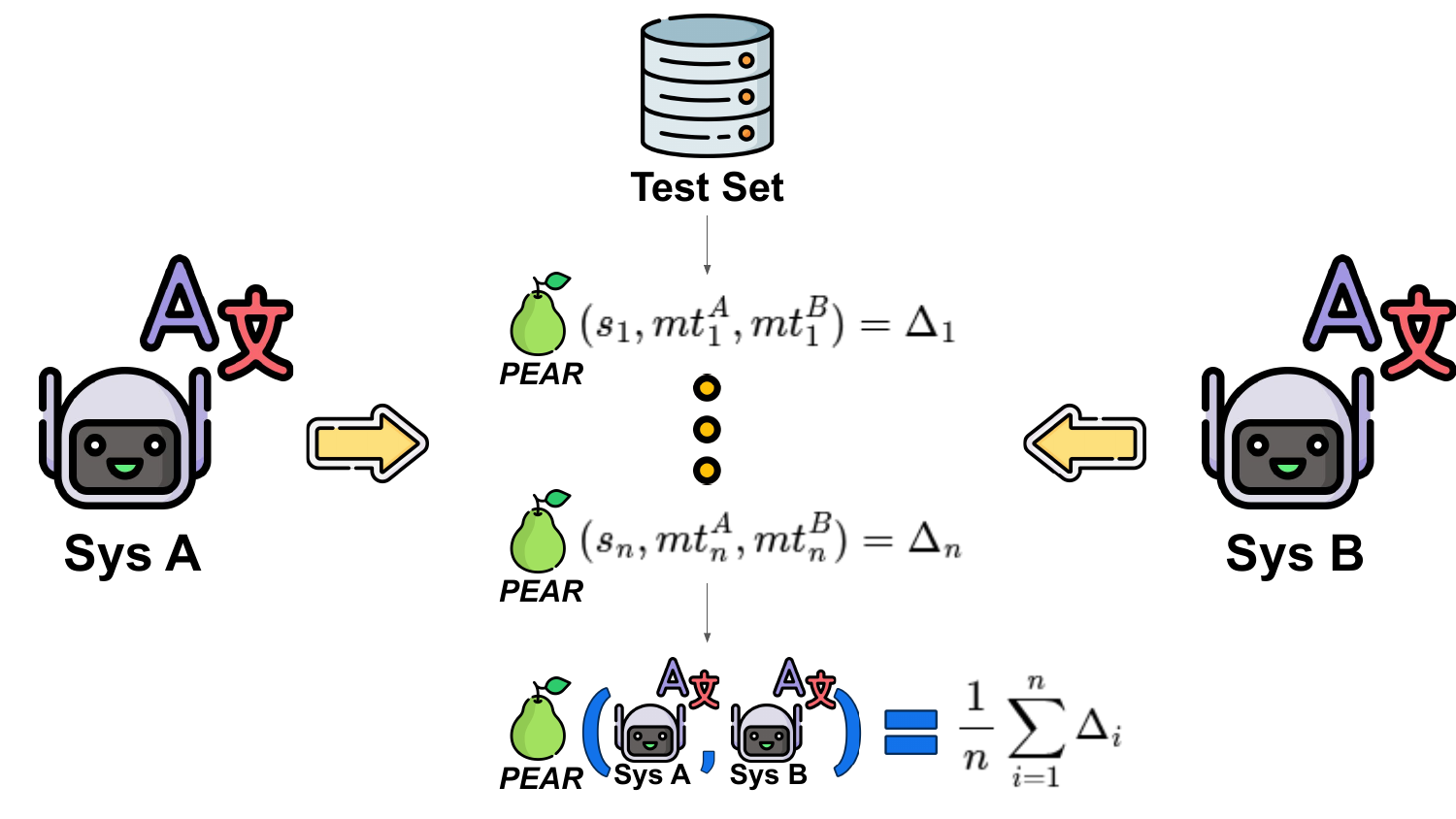}
  \caption{Segment- and system-level \pearfamily evaluation. For each source segment $s_i$, \pearfamily compares the system outputs $(mt_i^A, mt_i^B)$ and predicts a relative score $\Delta_i$. The sign indicates which translation is preferred ($\Delta_i>0$: $mt_i^A$; $\Delta_i<0$: $mt_i^B$), while the magnitude $|\Delta_i|$ reflects the strength of the preference, with smaller magnitudes corresponding to weaker preferences approaching a tie. The system-level \pearfamily score is then the arithmetic average of the segment-level \pearfamily scores.}
  \label{fig:pear-draw}
  \vspace{-2pt} 
\end{figure}

%% file: tables/pear_vs_single_1col.tex
\begin{table}[t]
\centering
\small
\setlength{\tabcolsep}{4pt}
\resizebox{\columnwidth}{!}{%
\begin{tabular}{l r c r r c}
\toprule
\textbf{Metric} & \textbf{$\boldsymbol{\theta}$} & \textbf{SPA} & \textbf{acc$_{eq}^*$} & \textbf{Avg Corr} & \textbf{Stat Sign Rank} \\
\midrule

Single-QE-XL$_{\mathrm{KD}}$ & 3.5B  & 80.9 & 57.9 & 69.4 & 2 \\
Single-QE-XL                 & 3.5B  & 80.4 & 57.6 & 69.0 & 3 \\
Single-QE$_{\mathrm{KD}}$    & 560M  & 80.6 & 57.4 & 69.0 & 3 \\
Single-QE                    & 560M  & 80.0 & 57.2 & 68.6 & 4 \\
\midrule

PEAR-XL$_{\mathrm{both,KD}}$ & 3.5B  & \textbf{82.1} & 58.1 & \textbf{70.1} & \textbf{1} \\
PEAR-XL$_{\mathrm{KD}}$      & 3.5B  & 82.0 & \textbf{58.2} & \textbf{70.1} & \textbf{1} \\
PEAR-XL$_{\mathrm{both}}$    & 3.5B  & 81.4 & \textbf{58.2} & 69.8 & \textbf{1} \\
PEAR-XL                      & 3.5B  & 81.5 & 58.1 & 69.8 & \textbf{1} \\
PEAR$_{\mathrm{both,KD}}$    & 560M  & 81.9 & 58.1 & 70.0 & \textbf{1} \\
PEAR$_{\mathrm{KD}}$         & 560M  & 81.8 & \textbf{58.2} & 70.0 & \textbf{1} \\
PEAR$_{\mathrm{both}}$       & 560M  & 81.2 & 58.0 & 69.6 & 2 \\
PEAR                         & 560M  & 80.9 & 57.9 & 69.4 & 2 \\
\bottomrule
\end{tabular}%
}

\caption{Controlled comparison of PEAR against matched single-candidate QE baselines on the WMT24 MQM test set.
SPA and acc$_{eq}^*$ are averaged over English--German, English--Spanish, and Japanese--Chinese;
Avg Corr is the mean of these two averages.
Stat Sign Rank reports rankings computed with the PERM-BOTH hypothesis test \citep{deutsch-etal-2021-statistical}, following the WMT24 Metrics Shared Task setup \citep{freitag-etal-2024-llms}.
Bold indicates the best score in each column.
For PEAR, \textit{both} denotes the bidirectional pairwise QE configuration.
\textit{KD} denotes models fine-tuned with additional MQM supervision distilled from GPT-4.1-mini.}
\label{tab:pear_vs_single}
\end{table}

%% file: tables/wmt24_main_res.tex
\begin{table*}[t]
\centering
\small
\setlength{\tabcolsep}{4.5pt}

\begin{tabular*}{\textwidth}{@{\extracolsep{\fill}}%
    >{\centering\arraybackslash}p{0.10\textwidth}
    >{\raggedright\arraybackslash}p{0.34\textwidth}
    r c r r r r@{}}
\toprule
\textbf{Group} & \textbf{Metric} & \textbf{$\boldsymbol{\theta}$} & \textbf{Ref?} & \textbf{SPA} & \textbf{acc$_{eq}^*$} & \textbf{Avg Corr} & \textbf{Stat Sign Rank} \\
\midrule

\multirow{10}{0.10\textwidth}{\centering\textbf{WMT24\\Metrics}}
& MetricX-24-Hybrid-QE-XXL    & 13B   & $\times$ & 84.9 & 58.0 & \textbf{71.4} & \textbf{1} \\
& MetricX-24-Hybrid-QE-XL     & 3.7B  & $\times$ & 83.4 & 56.5 & 69.9 & 3 \\
& MetricX-24-Hybrid-QE-Large  & 1.2B  & $\times$ & 80.6 & 56.1 & 68.3 & 6 \\
& XCOMET-QE                   & \multicolumn{1}{r}{24B\rlap{$^{\ddagger}$}} & $\times$ & 83.3 & 55.7 & 69.5 & 4 \\
& CometKiwi-XXL               & 10.5B & $\times$ & \textbf{85.4} & 55.2 & 70.3 & 3 \\
& CometKiwi                   & 560M  & $\times$ & 73.3 & 54.7 & 64.0 & 7 \\
& GEMBA-ESA                   & GPT-4 & $\times$ & 84.6 & 57.6 & 71.1 & \textbf{1} \\
& MetricX-24-Hybrid-Large     & 1.2B  & $\checkmark$ & 84.0 & 57.0 & 70.5 & 2 \\
& COMET-22                    & 560M  & $\checkmark$ & 82.4 & 55.4 & 68.9 & 5 \\
& BLEURT-20                   & 579M  & $\checkmark$ & 82.1 & 55.0 & 68.6 & 5 \\

\cmidrule(l){2-8}

\multirow{8}{0.10\textwidth}{\centering\textbf{Ours}}
& PEAR-XL$_{\mathrm{ref,KD}}$   & 3.5B & $\checkmark^{\dagger}$ & 82.7 & 58.9 & 70.8 & \textbf{1} \\
& PEAR-XL$_{\mathrm{both,KD}}$  & 3.5B & $\times$               & 82.7 & 58.3 & 70.5 & 2 \\
& PEAR-XL$_{\mathrm{ref}}$      & 3.5B & $\checkmark^{\dagger}$ & 81.6 & \textbf{59.2} & 70.4 & 2 \\
& PEAR-XL$_{\mathrm{both}}$     & 3.5B & $\times$               & 81.6 & 58.8 & 70.2 & 3 \\
& PEAR$_{\mathrm{ref,KD}}$      & 560M & $\checkmark^{\dagger}$ & 82.6 & 58.7 & 70.7 & 2 \\
& PEAR$_{\mathrm{both,KD}}$     & 560M & $\times$               & 82.0 & 58.8 & 70.4 & 2 \\
& PEAR$_{\mathrm{ref}}$         & 560M & $\checkmark^{\dagger}$ & 81.5 & 59.0 & 70.3 & 3 \\
& PEAR$_{\mathrm{both}}$        & 560M & $\times$               & 81.2 & 59.0 & 70.1 & 3 \\

\bottomrule
\end{tabular*}

\caption{WMT24 MQM meta-evaluation. SPA and acc$_{eq}^*$ are averaged over English--German, English--Spanish, and Japanese--Chinese;
Avg Corr is the mean of these two averages.
Stat Sign Rank reports rankings computed with the PERM-BOTH hypothesis test \citep{deutsch-etal-2021-statistical}, following the WMT24 Metrics Shared Task setup \citep{freitag-etal-2024-llms}.
Bold indicates the best score in each column.
$\ddagger$ denotes an ensemble that combines two 10.7B checkpoints and one 3.5B checkpoint.
For PEAR, \textit{ref} ($\dagger$) denotes the reference-anchored configuration, \textit{both} denotes the bidirectional pairwise QE configuration, and \textit{KD} denotes models fine-tuned with additional MQM supervision distilled from GPT-4.1-mini.}
\label{tab:wmt24_main_res}
\vspace{-0.5em}
\end{table*}

%% file: tables/mtranker_comparison.tex
\begin{table}[t]
\centering
\small
\setlength{\tabcolsep}{3.5pt}
\resizebox{\columnwidth}{!}{%
\begin{tabular}{lccccc}
\toprule
\textbf{Metric} & \textbf{$\boldsymbol{\theta}$} & \textbf{En-De} & \textbf{En-Es} & \textbf{Ja-Zh} & \textbf{Avg Pair Acc} \\
\midrule
PEAR$_{\mathrm{KD}}$ & 560M & \textbf{69.7} & \textbf{68.9} & \textbf{68.2} & \textbf{68.9} \\
PEAR & 560M & 69.2 & 67.5 & 67.2 & 68.0 \\
MT-RANKER-XXL & 5.7B & 64.7 & 65.0 & 67.8 & 65.8 \\
\bottomrule
\end{tabular}%
}
\caption{Comparison with MT-RANKER-XXL on the subset of the WMT24 MQM test set obtained after removing translation pairs with tied human MQM assessments.}
\label{tab:mtranker_comparison}
\end{table}

%% file: figures/pearref_anchor_ranks.tex
\begin{figure}[t]
  \centering
  \includegraphics[width=\columnwidth]{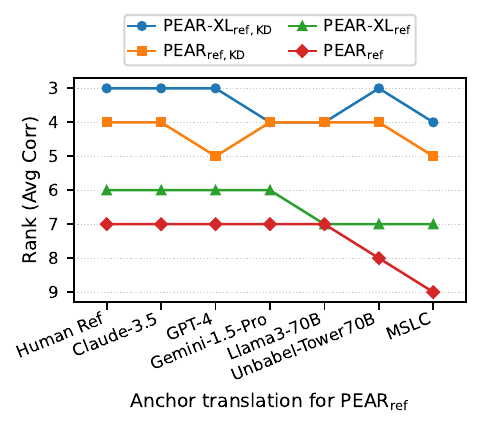}
  \vspace{-2em}
  \caption{Rank stability of PEAR$_{\mathrm{ref}}$ across anchors. The leftmost anchor (\textit{Human Ref}) uses the human reference as the fixed comparison translation, matching Table~\ref{tab:wmt24_main_res}. The remaining anchors use an MT output; for each MT anchor, that system is removed from the benchmark before recomputing meta-evaluation. Ranks are computed by Avg Corr (lower is better).}
  \label{fig:pearref_anchor_ranks}
\end{figure}

%% file: tables/mbr.tex
\begin{table}[t]
\centering
\scriptsize
\setlength{\tabcolsep}{4pt}

\begin{tabular}{@{}l l c c c@{}}
\toprule
\textbf{Utility} & \textbf{LP} & \textbf{XCOMET-XL} & \textbf{CometKiwi-XL} & \textbf{MetricX-XL} \\
\midrule
PEAR (full)     & En-De & \textbf{0.855} & \textbf{0.731} & \textbf{-5.2} \\
PEAR (sym.)     & En-De & 0.854 & \textbf{0.731} & -5.4 \\
COMET-22        & En-De & 0.844 & 0.730 & -6.6 \\
BLEURT-20       & En-De & 0.842 & 0.728 & -6.3 \\
\addlinespace
PEAR (full)     & En-Ja & \textbf{0.810} & \textbf{0.685} & \textbf{-6.4} \\
PEAR (sym.)     & En-Ja & 0.809 & \textbf{0.685} & -6.7 \\
COMET-22        & En-Ja & 0.798 & 0.684 & -7.8 \\
BLEURT-20       & En-Ja & 0.796 & 0.683 & -6.9 \\
\bottomrule
\end{tabular}

\caption{MBR decoding on 100-best lists for WMT24 En-De and En-Ja. PEAR (sym.) computes only one triangle of the $N\times N$ utility matrix (with $N$ candidates; here $N\!=\!100$) and fills the remainder by antisymmetry.}
\label{tab:mbr_results}
\end{table}

%% file: appendix.tex
\section{Pairwise Consistency of PEAR: Antisymmetry and Transitivity}
\label{app:pear_consistency}

PEAR is trained to assess graded quality differences between pairs of translations
(Section~\ref{sec:pear}).
We focus on two consistency properties of relative scoring: \emph{antisymmetry} and \emph{transitivity}.
Antisymmetry captures sign inversion under candidate reversal, whereas transitivity constrains score composition over triples.
In this section, we quantify PEAR's adherence to these two consistency properties and how the antisymmetry auxiliary training objective (Section~\ref{subsec:pear-objective}) affects both.

\paragraph{Setup.}
We run this analysis on the WMT24 MQM benchmark across English--German, English--Spanish, and Japanese--Chinese.
We focus on the most basic PEAR variant,\footnote{InfoXLM Large checkpoint.} trained without distilled MQM data
and evaluated in the single-pass inference configuration.
We compare two checkpoints that differ only in the antisymmetry regularization strength:
$\lambda_{\text{flip}} = 0$ and $\lambda_{\text{flip}} = 0.1$.

\paragraph{Antisymmetry deviation.}
Let $\Delta(S_A, S_B)$ denote the system-level PEAR score for the ordered system pair $(S_A, S_B)$,
computed as in Section~\ref{subsec:pear-formulation} by averaging segment-level scores. Antisymmetry requires
$\Delta(S_A, S_B) = -\Delta(S_B, S_A)$.
We measure deviation from this property using the absolute \emph{antisymmetry residual}:
\begin{equation*}
\label{eq:antisym_residual}
\epsilon_{\textsc{as}}(S_A, S_B) =
\bigl| \Delta(S_A, S_B) + \Delta(S_B, S_A) \bigr|.
\end{equation*}
We compute this over all MT system pairs.

\paragraph{Transitivity deviation.}
A stronger additive consistency property over triples is:
\begin{equation*}
\label{eq:transitivity}
\Delta(S_A, S_C) = \Delta(S_A, S_B) + \Delta(S_B, S_C).
\end{equation*}
We quantify deviation using the absolute \emph{transitivity residual}:
\begin{equation*}
\label{eq:trans_residual}
\begin{aligned}
\epsilon_{\textsc{tr}}(S_A, S_B, S_C)
= &\Bigl| \Delta(S_A, S_C) -\\
&(\Delta(S_A, S_B) + \Delta(S_B, S_C)) \Bigr|.
\end{aligned}
\end{equation*}
We compute this over all triples of distinct MT systems $(S_A, S_B, S_C)$.

\paragraph{Relative residual reporting.}
We report residuals relative to the typical magnitude of system-level PEAR scores.
This is preferable to raw residual values because adding or removing the antisymmetry regularization training objective could change the output score distribution, making absolute residuals less comparable across training setups.
Concretely, we normalize by the mean absolute system-level score:
\begin{equation*}
\label{eq:mean_abs_delta}
\mu_{\Delta} = \mathbb{E}_{(S_A,S_B)} \bigl[ \, |\Delta(S_A, S_B)| \, \bigr],
\end{equation*}
and report the relative deviations:
\begin{equation*}
\label{eq:relative_residuals}
\begin{aligned}
\rho_{\textsc{as}}
&= \frac{\mathbb{E}_{(S_A,S_B)}\bigl[\epsilon_{\textsc{as}}(S_A,S_B)\bigr]}{\mu_{\Delta}}, \\
\rho_{\textsc{tr}}
&= \frac{\mathbb{E}_{(S_A,S_B,S_C)}\bigl[\epsilon_{\textsc{tr}}(S_A,S_B,S_C)\bigr]}{\mu_{\Delta}}.
\end{aligned}
\end{equation*}

\paragraph{Results.}
Table~\ref{tab:pear_consistency} shows that enabling the antisymmetry auxiliary training objective substantially reduces antisymmetry deviations, yielding residuals that are only a small fraction of the typical PEAR score magnitude.
While transitivity deviations remain larger overall (as expected for an unconstrained real-valued scorer), they are also consistently reduced with $\lambda_{\text{flip}} = 0.1$.
This indicates that encouraging sign inversion under candidate reversal improves not only antisymmetry, but also promotes a more consistent behavior across MT system triples.

\input{tables/pear_antisymmetry_transitivity}

\section{Huber Loss vs.\ MSE for Pairwise Difference Regression}
\label{app:huber_vs_mse}

PEAR is trained to regress human quality differences $\Delta^\star_{ab}$ (Section~\ref{subsec:pear-objective}). In all the experiments, we use the Huber loss, which is less sensitive to occasional large residuals than MSE \citep{huber1964robust}.

To quantify the impact of this choice, we run an ablation for PEAR\footnote{We use the InfoXLM Large variant.} in the same training setup as Section~\ref{subsec:scale}, changing only the regression loss from Huber to MSE while keeping all other hyperparameters fixed. We evaluate on WMT24 MQM with the official meta-evaluation toolkit, reporting Soft Pairwise Accuracy (SPA) at the system level \citep{thompson-etal-2024-improving}, pairwise accuracy with tie calibration (acc$_{eq}^*$) at the segment level \citep{deutsch-etal-2023-ties}, and Avg Corr.

Table~\ref{tab:huber_vs_mse_wmt24} shows that Huber yields small but consistent gains over MSE across meta-evaluation statistics. We hypothesize that down-weighting large residuals stabilizes learning under heavy-tailed pairwise targets, which can improve calibration for close comparisons among strong MT systems. Further analysis of when and why these gains arise is left to future work.

\input{tables/huber_vs_mse}

\section{Training and Evaluation Setup}
\label{subsec:baselines}

\subsection{Data}

PEAR training set is composed of 7M translation pairs across 51 language pairs for the first training stage (DA and DA+SQM), and by 6M across 10 language pairs for the second training stage (MQM). We do not apply any transformation to the scores provided by human annotators. Following \citet{proietti-etal-2025-estimating}, when multiple MQM scores are available for a translation, we do not aggregate them and instead treat each annotation as a separate training instance. For the KD data, we run the GEMBA-MQM V2 approach on WMT data annotated only with DA and DA+SQM, and on internal MT output data, featuring a total of 2M additional translation pairs, covering six additional language pairs in addition to those present in gold MQM data.

\subsection{Hyperparameters}

For the experiments outlined in Section~\ref{subsec:scale}, we train all the models with AdamW \citep{loshchilov2018decoupled}, using a learning rate equal to 2e-5 until convergence on the development set (WMT23 MQM data) for a maximum of five epochs. For the experiments described in Section~\ref{subsec:wmt24_res}, we keep the same hyperparameters, extend the training set with WMT23, and train for one epoch. The $\lambda_{\text{flip}}$ hyperparameter has been set to 0.1 after preliminary hyperparameter optimization experiments where we did not observe large differences in the [0.1, 0.5] range, with a step size equal to 0.1. The $\delta$ hyperparameter has been set to 4.5 after preliminary hyperparameter optimization experiments, where we did not observe large differences in the [2.0, 8.0] range, with a step size equal to 0.5. Additionally, in our initial experiments (Section~\ref{subsec:scale}), including the antisymmetry auxiliary training objective yielded a higher Avg Corr on the overall WMT24 evaluation. Without it, PEAR's Avg Corr decreased from 69.4\footnote{This refers to the performance obtained by \pear in Table~\ref{tab:pear_vs_single}.} to 68.9. The parameter $\alpha_{\mathrm{raw}}$ presented in Section~\ref{sec:pear} is kept frozen for the first training stage on DA and DA+SQM data, and it is learned in the second training stage on MQM, with an initial value of 1.0. This is intended to facilitate adaptation to the training distribution in the second stage, which is derived from MQM annotations rather than DA or DA+SQM.

\subsection{Baselines}

Since PEAR targets improvements in QE for comparing candidate translations, we primarily benchmark against top-performing QE metric families submitted as primary entries to the WMT24 Metrics Shared Task \citep{freitag-etal-2024-llms}. For additional context, we also report a set of strong WMT24 reference-based baselines.

\subsubsection{QE metrics.}

We compare with the following metric families:

\begin{itemize}
    \item \textbf{COMET family.} We include CometKiwi \citep{rei-etal-2022-cometkiwi}, its large-scale variant CometKiwi-XXL \citep{rei-etal-2023-scaling}, and XCOMET-QE \citep{guerreiro-etal-2024-xcomet}.
    \item \textbf{MetricX family.} We include the QE variants of MetricX-24 Hybrid \citep{juraska-etal-2024-metricx}, namely MetricX-24-Hybrid-QE-Large, -XL, and -XXL.
    \item \textbf{GEMBA family.} We include GEMBA-ESA, a WMT24 primary submission that prompts GPT-4 to follow the ESA procedure by extracting error spans and then producing a final 0--100 score \citep{kocmi-federmann-2023-gemba,kocmi-federmann-2023-large,freitag-etal-2024-llms}.
\end{itemize}

We do not include COMET-poly in the WMT24 comparison: the publicly released checkpoints are trained on WMT data up to 2024 \citep{zufle-etal-2025-comet}, which overlaps with the benchmark.

\subsubsection{Reference-Based Metrics}
To provide an additional performance reference from strong reference-based metrics, we include COMET-22 \citep{rei-etal-2020-comet,freitag-etal-2024-llms}, BLEURT-20 \citep{sellam-etal-2020-learning}, and MetricX-24-Hybrid-Large \citep{juraska-etal-2024-metricx}.
These models score single candidate translations in a reference-based setting and serve as competitive anchors alongside the comparisons focused on QE.

\section{Correlation Matrices}
\label{app:corr_mats}

\input{figures/corrs_en-de}
\input{figures/corrs_en-es}
\input{figures/corrs_ja-zh}

This appendix section provides the correlation matrices for the WMT24 MQM language pairs En-De (Figure~\ref{fig:pear_corr_en_de}), En-Es (Figure~\ref{fig:pear_corr_en_es}), and Ja-Zh (Figure~\ref{fig:pear_corr_ja_zh}), which are discussed in Section~\ref{subsec:pear_metric_corr}.

%% file: tables/pear_antisymmetry_transitivity.tex
\begin{table}[t]
\centering
\small
\begin{tabular}{lcc}
\toprule
Model (InfoXLM Large) & $\rho_{\textsc{as}} \downarrow$ & $\rho_{\textsc{tr}} \downarrow$ \\
\midrule
PEAR ($\lambda_{\text{flip}} = 0$)   & 0.196 & 0.561 \\
PEAR ($\lambda_{\text{flip}} = 0.1$)   & \textbf{0.014} & \textbf{0.189} \\
\bottomrule
\end{tabular}
\caption{
Pairwise consistency deviations on WMT24 MQM (En-De, En-Es, Ja-Zh), computed over all system pairs (antisymmetry) and system triples (transitivity). Residuals are reported as proportions relative to the mean absolute system-level PEAR score, i.e.,~lower is better.}
\label{tab:pear_consistency}
\end{table}

%% file: tables/huber_vs_mse.tex
\begin{table}[t]
\centering
\small
\begin{tabular}{lccc}
\toprule
\textbf{Loss} & \textbf{SPA} $\uparrow$ & \textbf{acc$_{eq}^*$} $\uparrow$ & \textbf{Avg Corr} $\uparrow$ \\
\midrule
MSE   & 80.6 & 57.6 & 69.1 \\
Huber & \textbf{80.9} & \textbf{57.9} & \textbf{69.4} \\
\bottomrule
\end{tabular}
\caption{
WMT24 meta-evaluation for PEAR (InfoXLM Large, 560M) trained with MSE vs.\ Huber regression on pairwise human-difference supervision.
SPA is reported at the system level \citep{thompson-etal-2024-improving}; acc$_{eq}^*$ is segment-level pairwise accuracy with tie calibration \citep{deutsch-etal-2023-ties}.
}
\label{tab:huber_vs_mse_wmt24}
\end{table}

%% file: figures/corrs_en-de.tex
\begin{figure*}[t]
  \centering
  \includegraphics[width=0.95\textwidth]{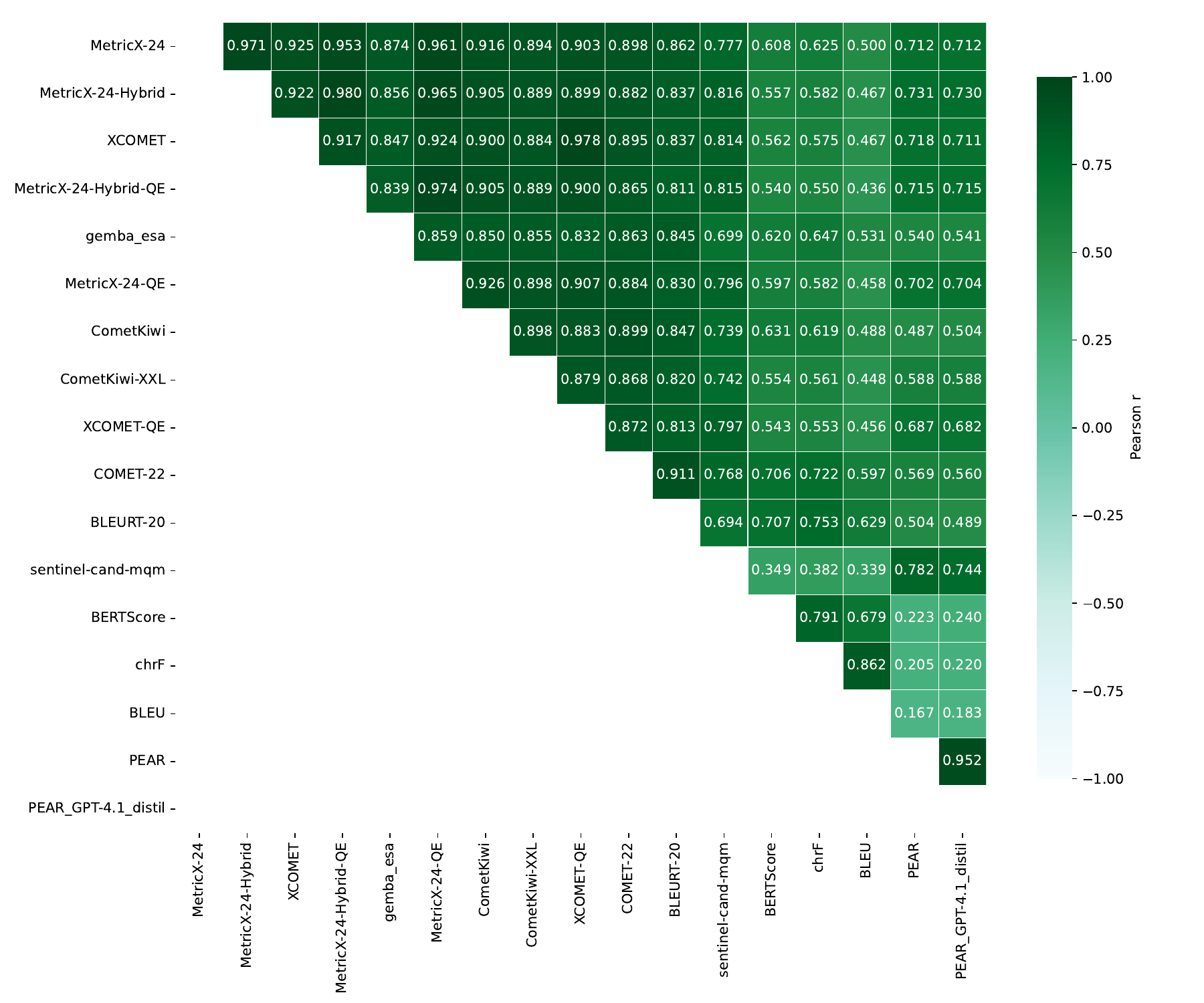}
  \caption{En-De Pearson correlation matrix between segment-level pairwise difference scores derived from WMT24 metric scores. Single-candidate metrics are converted into pairwise differences by subtraction, while PEAR produces pairwise scores directly. The two PEAR models are instantiated with InfoXLM Large and used in the bidirectional (\textit{both}) inference mode.}
  \label{fig:pear_corr_en_de}
\end{figure*}

%% file: figures/corrs_en-es.tex
\begin{figure*}[t]
  \centering
  \includegraphics[width=0.95\textwidth]{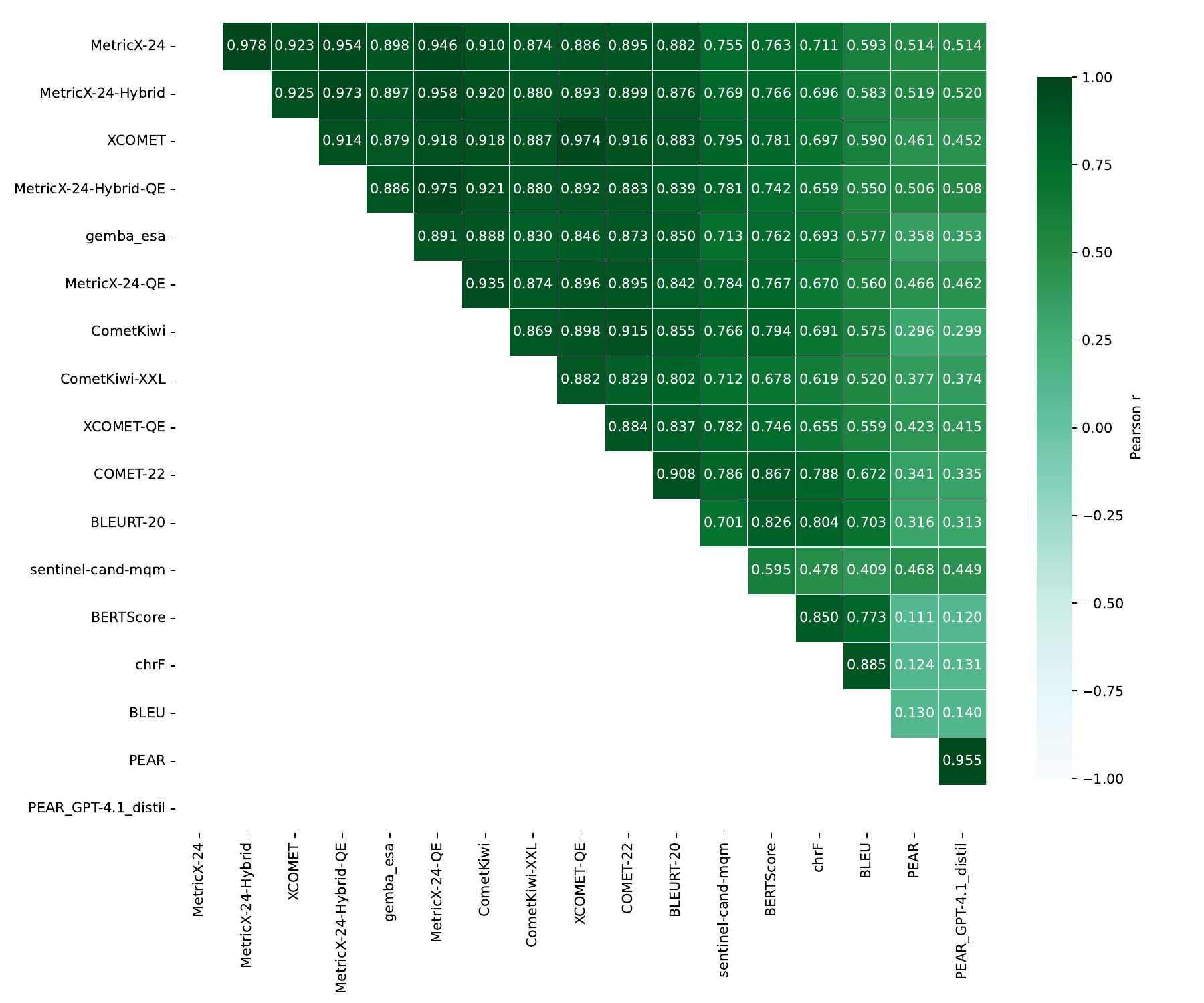}
  \caption{En-Es Pearson correlation matrix between segment-level pairwise difference scores derived from WMT24 metric scores. Single-candidate metrics are converted into pairwise differences by subtraction, while PEAR produces pairwise scores directly. The two PEAR models are instantiated with InfoXLM Large and used in the bidirectional (\textit{both}) inference mode.}
  \label{fig:pear_corr_en_es}
\end{figure*}

%% file: figures/corrs_ja-zh.tex
\begin{figure*}[t]
  \centering
  \includegraphics[width=0.95\textwidth]{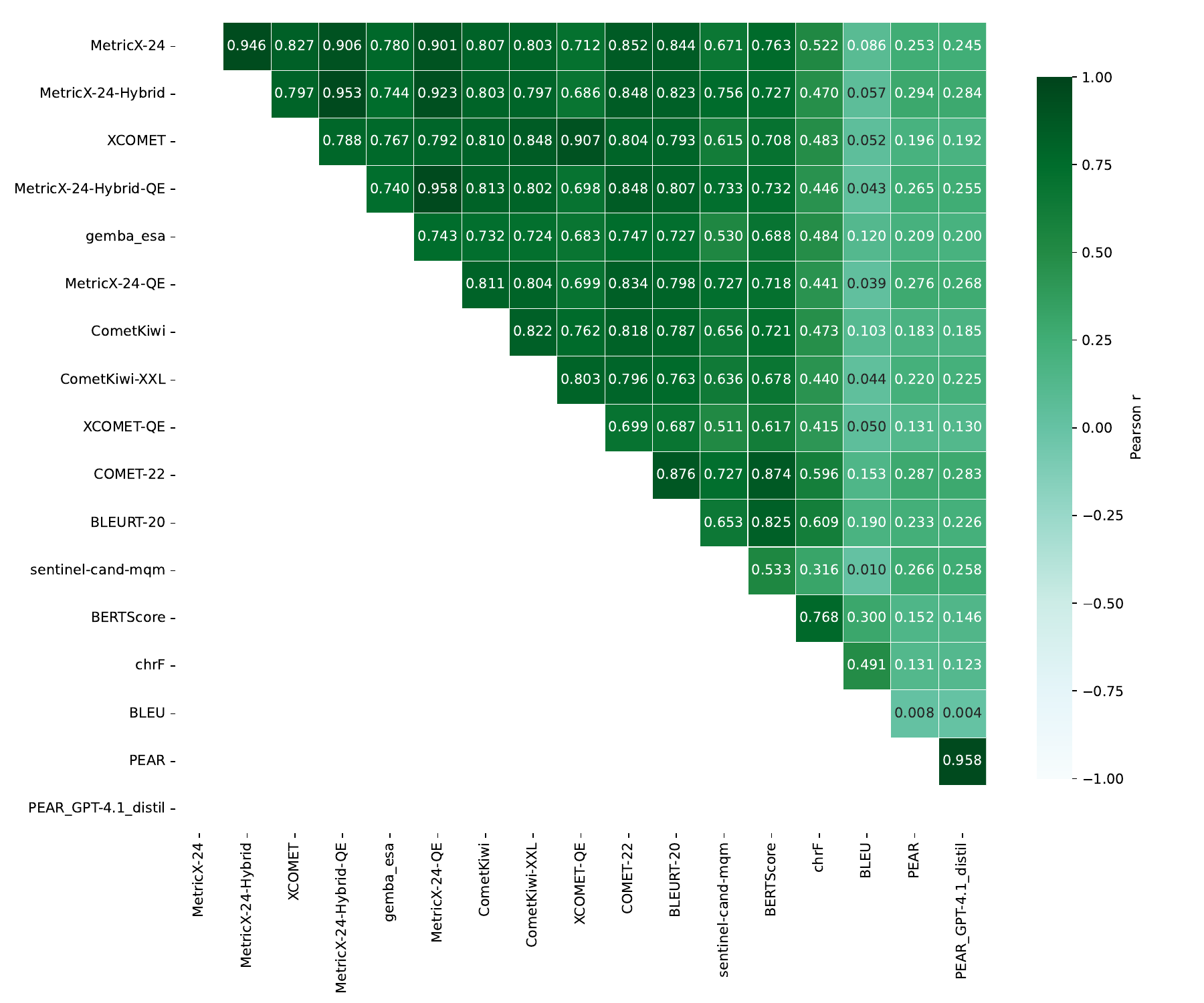}
  \caption{Ja-Zh Pearson correlation matrix between segment-level pairwise difference scores derived from WMT24 metric scores. Single-candidate metrics are converted into pairwise differences by subtraction, while PEAR produces pairwise scores directly. The two PEAR models are instantiated with InfoXLM Large and used in the bidirectional (\textit{both}) inference mode.}
  \label{fig:pear_corr_ja_zh}
\end{figure*}